\documentclass[letterpaper, 10 pt, conference]{ieeeconf}

\IEEEoverridecommandlockouts

\usepackage[letterpaper,
            left=0.765in,
            right=0.78in,
            top=1.0in,
            bottom=0.79in]{geometry}
\usepackage{amssymb}
\usepackage{amsmath}
\usepackage{amsfonts}
\usepackage{listings}
\usepackage{hyperref}
\usepackage{graphicx}
\usepackage{wrapfig}
\usepackage{xcolor}
\usepackage{mathtools}
\usepackage{algorithm}
\usepackage{algpseudocode}
\usepackage{textcomp}
\usepackage{makecell}

\usepackage{microtype}
\usepackage[subtle]{savetrees}
\newtheorem{theorem}{Theorem}[section]
\newtheorem{definition}{Definition}[section]

\definecolor{pink}{rgb}{0.8, 0.2, 0.5}

\title{\LARGE \bf
Localization in Dynamic Planar Environments Using Few Distance Measurements  
}

\author{Michael M.~Bilevich$^{1}$ \and Shahar Guini$^{1}$ \and Dan Halperin$^{1}$%
    \thanks{$^{1}$Blavatnik School of Computer Science, Tel-Aviv University, Israel. Work by M.B.\ and D.H.\  has been supported in part by the Israel Science
Foundation (grant no.~2261/23), 
by NSF/US-Israel-BSF (grant no.~2019754),
by the Blavatnik Computer Science Research Fund, 
and by the Shlomo Shmelzer Institute for 
Smart Transportation at Tel Aviv University.}%
}%

\begin{document}

\maketitle
\thispagestyle{empty}
\pagestyle{empty}

\begin{abstract}

We present a method for determining the unknown location of a sensor placed in a known 2D environment in the presence of unknown dynamic obstacles, using only few distance measurements. 
We present guarantees on the quality of the localization, which are robust under mild assumptions on the density of the unknown/dynamic obstacles in the known environment.
We demonstrate the effectiveness of our method in simulated experiments for different environments and varying dynamic-obstacle density. Our open source software is available at \href{https://github.com/TAU-CGL/vb-fdml2-public}{https://github.com/TAU-CGL/vb-fdml2-public}.

\end{abstract}

\section{Introduction}

Robot localization is the task of determining the pose (or location) of a robot in some environment, and is an extensively researched problem in  robotics~\cite{skrzypczynski2017, PANIGRAHI20226019, pak2023state}. The localization can be carried out with various sensors and techniques, and in different environments. In this work we focus on the ``kidnapped robot'' variant~\cite{pak2023state}, which strives to find the robot's location in an environment, with no prior information on its location, as opposed to fine-tuning the localization assuming we know generally where the robot is.

In a previous work~\cite{bilevich23}, we presented a method for performing robot localization in a planar (known) environment with only a few distance-measurements. 
However, environments may also contain dynamic disturbances, which do not appear in the known map of the environment, such as moved furniture, people walking around, other robots, etc. In this work, we present a general method for few distance-measurement localization, which is robust to such dynamic changes, both in theory and experiments.

\section{Problem Statement}
\label{problem-statement}

The sensor is placed in the interior of a planar workspace $\mathcal{W}\subseteq \mathbb{R}^2$.
Let $\mathcal{D}_1,\dots,\mathcal{D}_m \subseteq \mathcal{W} \subset \mathbb{R}^2$ be closed planar regions which are the dynamic obstacles, with corresponding trajectories $\varphi_1,\dots,\varphi_m:\mathbb{R}_{\geq 0}\to SE(2)$, such that at time $t\geq 0$, the free region becomes 
\[
\mathcal{W}_t \coloneqq \mathcal{W} \setminus \left(\bigcup_{i=1}^m \varphi_i(t)\left(\mathcal{D}_i\right)\right)\;.
\]

We refer to $\mathcal{W}$ as the \emph{static} environment, and to $\mathcal{W}_t$ as the current (at time $t$) \emph{dynamic} environment. 

A distance measurement is a mapping
\begin{align}
h: \mathbb{R}_{\geq 0} \times (\mathcal{W}\times \mathbb{S}^1) \subset \mathbb{R}_{\geq 0} \times SE(2) \rightarrow \mathbb{R}_{\geq 0} ,
\end{align}
such that $h(t, x,y,\theta)$ is the length of the 
first intersection of a ray emanating from $(x,y)$ in direction $\theta$ with the boundary of the 
workspace $\mathcal{W}_t$ at time $t$. 

Furthermore, denote by $h\big|_\mathcal{W}:\mathcal{W}\times \mathbb{S}^1\rightarrow \mathbb{R}_{\geq 0}$ the distance measurement for the known workspace $\mathcal{W}$, without dynamic obstacles, defined in the same way as $h$.
We are now ready to state the basic version of the problem that we study.

\smallskip

\noindent\emph{The problem:}
Given a static workspace $\mathcal{W}$, 
dynamic obstacles $\mathcal{D}_1,\dots,\mathcal{D}_m$
with corresponding trajectories
$\varphi_1,\dots,\varphi_m$ which are unknown, 
a set $g_1,\ldots,g_k$ of rigid-body transformations with a set $d_1, \ldots, d_k$ of corresponding positive real values (which were taken at times $t_1, \dots, t_k$, respectively), 
we wish to find all the poses $(x,y,\theta)$ such that $(x,y)\in \mathcal{W}$ and $d_i=h(t_i, g_i(x,y,\theta))$ for all $i=1,\dots,k$.

\medskip

We wish to find the configuration $(x,y,\theta)$, which is the original pose of the robot. 
Before each one of the $k$ measurements, the robot moves to another pose or stays put. The pose of the sensor when making the $i$th distance measurement at time $t_i$ is $g_i(x,y,\theta)$.

\begin{figure}[t!]
    \centering
    \includegraphics[width=0.33\textwidth]{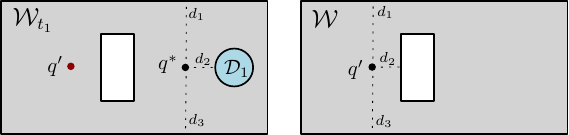}
    \caption{
    Example for why sampling a dynamic obstacle might lose the ground truth location. Our workspace $\mathcal{W}$ is in gray. We have one dynamic obstacle $\mathcal{D}_1$ with $\varphi_1 = 1 \in SE(2)$ (see remark in Section~\ref{problem-statement}).
    We take three distance measurements $d_i$ with a rotation offset of $\pi/2$ radians clockwise. The robot is at $q_* \in SE(2)$.
    Left: The free region at time $t_1$, $\mathcal{W}_{t_1}$. Ignoring the existence of dynamic obstacle yields the red pose $q'$ which is false.
    Right: When we ignore the dynamic obstacles, and look for locations for which we would measure $d_i$, we get only $q'$ and lose $q_*$.
    }
    \label{fig:incorrect}
\end{figure}

As shown in~\cite{bilevich23}, localization in a completely known environment can be effectively approximated. 
If, a-priori we know exactly how $\mathcal{W}_{t_i}$ looks like, then that  method~\cite{bilevich23} would work as-is.
Obviously, in the presence of unknown obstacles, it could not be applied as-is. 
See Figure~\ref{fig:incorrect} for an example.

We focus here on workspaces that have a polygonal boundary, namely polygons or polygons with holes. Aiming for generality of the approach, we assume no prior knowledge on the topology, geometry, number or location of the dynamic obstacles. Of course we must make some assumptions on the dynamic obstacles in order for the problem to be solvable. Indeed, we make the following \emph{$(k,k')$-dynamic sparsity assumption}: We specify two natural numbers $3\leq k' \leq k$ and assume that out of a batch of $k$ measurements, there would be at least $k'$ which measure the distance from the boundary of the known workspace $\mathcal{W}$.

\medskip
\noindent\textbf{Remark.}
A trajectory $\varphi_i$ of a dynamic obstacle may be degenerate, in the sense that $\varphi_i = 1 \in SE(2)$, i.e., for all $t\geq 0$ the obstacle stays put. 
For simplicity, we refer to such unknown obstacles as dynamic as well.


\section{The method}
\label{sec:method}

For any $a\in \mathbb{N}$, denote $[a] = \{1,\dots,a\}$. Assume that the robot has measured distances $d_1,\dots,d_k$ for $k \geq 4$, and assume that there is  subset of size $k'$ of measurements, $3 \leq k' \leq k$, which sample the static environment. 
Let $M_{d_i}$ be the preimage of the distance measurement in $\mathcal{W}$,
\begin{align}
    M_{d_i} \coloneqq \left(h\big|_{\mathcal{W}} \circ g_i \right)^{-1}
    (\{d_i\})\;,
\end{align}
and let $\hat{M}_{d_i}$ be the corresponding voxel clouds approximations (of resolution $n\times n \times n$, for a given resolution parameter $n$) of those preimages $M_{d_i}$ for $i\in [k]$. 
We get $\hat{M}_{d_i}$ from the method described in~\cite{bilevich23}.

We also define 
\emph{agreement} of a measurement as follows:

\begin{definition}
    Let $\varepsilon > 0$. For a measurement $d_\alpha$, with $\alpha \in [k]$, we say that a pose $q_j$ \emph{$\varepsilon$-agrees} (with the static environment) on the measurement $d_\alpha$ if 
    
    \begin{align}
        \left|\left(h\big|_\mathcal{W} \circ g_\alpha\right)(q_j) - d_\alpha \right| < \varepsilon\;.
    \end{align}
\end{definition}

\bigskip

Then we compute (where $2^{[k]}$ is the collection of all subsets $\sigma$ of $[k]$):

\begin{align}
    \hat{M}_{k'} = \bigcup_{\substack{\sigma \in 2^{[k]} \\ |\sigma|=k'}} 
    \bigcap_{\alpha \in \sigma} \hat{M}_{d_\alpha}\;.
\end{align}

\smallskip

\begin{theorem}
\label{thm:main}
    The voxel cloud approximation $\hat{M}_{k'}$ is \emph{conservative}\footnote{Up to some small set of voxels $\mathcal{A}_{n,\mathcal{W}}$, which we can treat specifically.}. That is, if $q_* \in SE(2)$ is the ground truth, then $q_* \in \hat{M}_{k'}$. Furthermore, the distance between $q_*$ and the nearest predicted localization in $\hat{M}_{k'}$ is $O(1/n)$.
\end{theorem}
        
\smallskip

We then extract a collection $\{q_j\}_j$ 
of poses which are the centers of mass of connected components of voxels in $\hat{M}_{k'}$. 
However, this set might contain many irrelevant poses, which are far from representing the correct localization. Hence we only leave poses $q_j$ that fulfill the following conditions, for prescribed parameters $\delta,\varepsilon>0$:

\begin{itemize}
    \item The pose $q_j$ is unique: $\forall j'\neq j$, $||q_j - q_{j'} || \geq \delta$.
    
    \item 
    The pose $q_j$ $\varepsilon$-agrees with $d_\alpha$ for every $\alpha \in \sigma$. 
\end{itemize}


\section{Experiments and Results}

Our code is written in C++ with Python Bindings. We utilize 
OpenMP\footnote{\url{https://www.openmp.org/}}
for parallel computation. The code is run on an Ubuntu machine with Intel Core i7-12700 CPU.

We demonstrate our performance on four different test scenes: A square room, a polygon based on a LiDAR scan of our lab, a floor-plan,
and randomly generated polygons.

\begin{figure} [t!]
    \centering
    \includegraphics[width=0.16\textwidth]{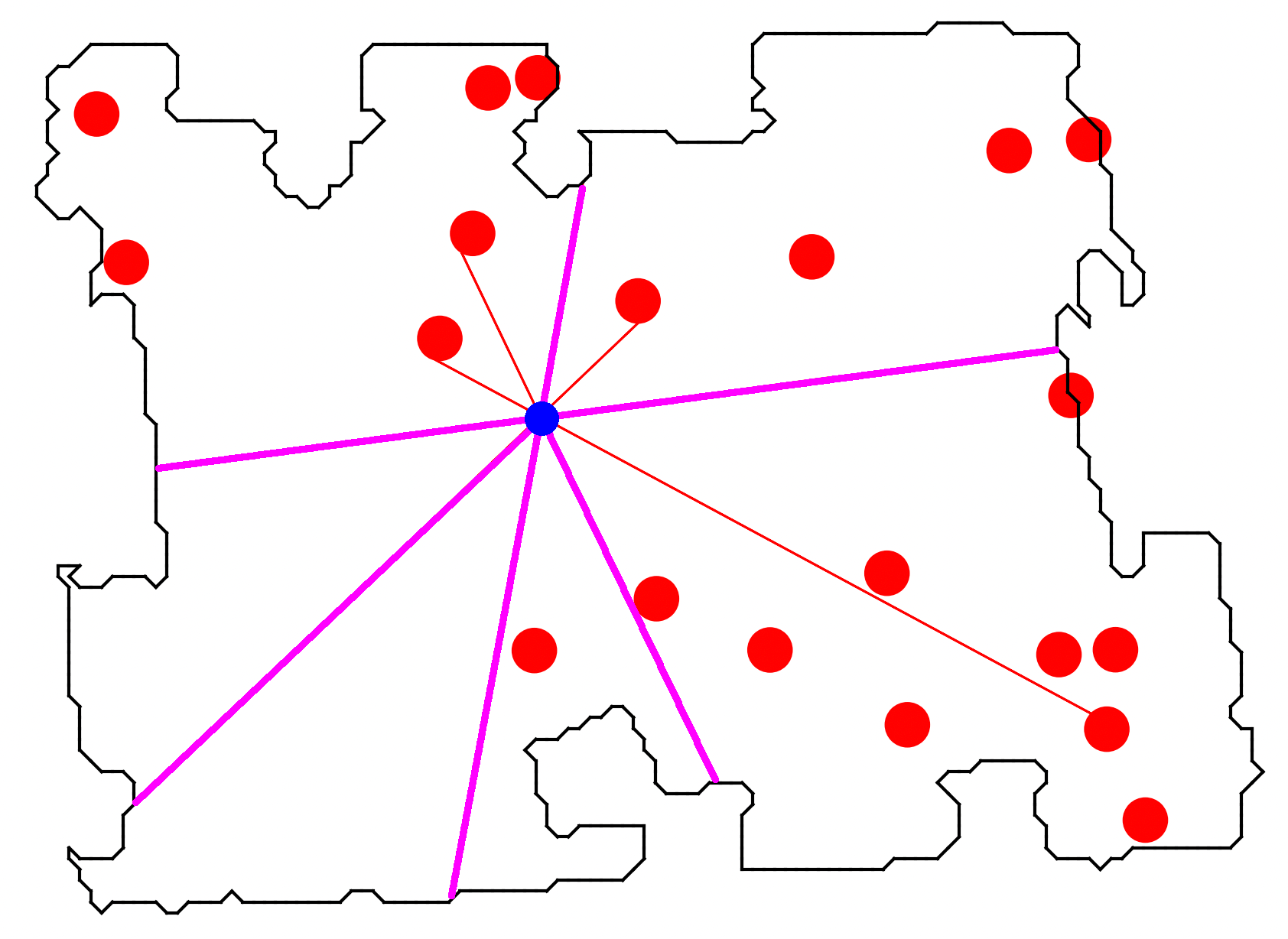}
    \caption{Example of the simulated experiment on the \texttt{lab-lidar} polygon, with $20$ dynamic obstacles. Ground truth location is in blue, and we cast $10$ rays for distance measurements, with $4$ of them sampling the dynamic obstacles (in red), and the rest sampling the static workspace $\mathcal{W}$ (in magenta).}
    \label{fig:measurements}
\end{figure}

Our experiments are carried out as follows: We randomly place the sensor in each of the aforementioned rooms, with randomly placed dynamic obstacles which stay in place (see remark at the end of Section~\ref{problem-statement}, and Figure~\ref{fig:measurements} for an example) and perform $k=10$ distance measurements, with $2\pi/k$ increments in rotation between every pair of consecutive measurements. We apply our method assuming (10,6)-dynamic sparsity, which does not always occur in our experiments. We repeat each experiment $50$ times for different grid resolutions, and for $m=10$ and $30$ dynamic obstacles. We also ran our base method on those scenarios. In Table~\ref{tab:success-rate} we indeed see that our method significantly improves the success rate.

\begin{table}[t!]
    \centering
    \begin{tabular}{|c||c|c|c|c|c|c|}
    \hline
    & \multicolumn{2}{|c|}{\texttt{lab-lidar}} & \multicolumn{2}{|c|}{\texttt{floor-plan}} & 
    \multicolumn{2}{|c|}{\texttt{random}} \\
    \hline
    $m$ & 10 & 30 & 10  & 30 & 10 & 30 \\
    \hline
    \hline

    FDML~\cite{bilevich23} & 32.3 & 18.6 & 14.0 & 13.4 & 34.9 & 29.1\\

    \hline
    
    \makecell{\textbf{Current}\\ \textbf{method}} & \textbf{94.3} & \textbf{92.4} & \textbf{94.6} & \textbf{88.6} & \textbf{99.1} & \textbf{93.1}\\
    \hline
    
    \end{tabular}
    \caption{Average success rate (\%) comparison for each scene.}
    \label{tab:success-rate}
\end{table}

\vspace{-1ex}

\section{Conclusion and Future Work}

In this work we showed that the few distance-measurement localization technique can be adjusted for uncertain obstacles in a known environment, and have demonstrated a significant improvement in performance on such scenarios.
Many details of the analysis and experiments have been omitted here and will be supplied in a forthcoming full version. 

However, we are yet to determine with high confidence which of the measurements are those that measure the static environment. Furthermore, we do not guarantee the dynamic sparsity of a given scenario (even if we have full information on the dynamic obstacles). 
The next goals are: (i) devise analysis tools for determining the dynamic sparsity of a given setting, and (ii) estimate the actual dynamic sparsity value in the absence of knowledge about the dynamic setting.

%
\bibliographystyle{IEEEtran}
\bibliography{bibliography}

\end{document}